# An optimal fuzzy-PI force/motion controller to increase industrial robot autonomy

Nuno Mendes, Pedro Neto, J. Norberto Pires and Altino Loureiro

*Department of Mechanical Engineering (CEMUC)-POLO II, University of Coimbra, Rua Luís Reis Santos, 3030-788 Coimbra, Portugal*

Tel.: 00351 239 790 700

Fax: 00351 239 790 701

Email: nuno.mendes@dem.uc.pt

**Abstract:** This paper presents a method for robot self-recognition and self-adaptation through the analysis of the contact between the robot end-effector and its surrounding environment. Often, in off-line robot programming the idealized robotic environment (the virtual one) does not reflect accurately the real one. In this situation we are in the presence of a Partially Unknown Environment (PUE). Thus, robotic systems must have some degree of autonomy to overcome this situation, especially when contact exists. The proposed force/motion control system has an external control loop based on forces and torques exerted on the robot end-effector and an internal control loop based on robot motion. The external control loop is tested with an optimal Proportional Integrative (PI) and a fuzzy-PI controller. The system performance is validated with real-world experiments involving contact in PUEs.

## 1. Introduction

Today, the robotics market imposes that robots are programmed more quickly, more easily and used in more challenging tasks [1-2]. In this context, off-line robot programming (OLP) is often considered a good solution. Different approaches to OLP have been proposed, most of them based on CAD data. This includes an OLP system based on a CAD/CAM/CAE software for the shoes manufacturing industry [3], the definition of robot paths for spray painting processes [4], the generation of robot paths from CAD for a friction stir welding process [5] and direct OLP from a common CAD package [6-8]. In addition, there are also OLP commercial software packages in which CAD drawings serve as its input. The problem is that all data from CAD (the CAD drawings representing the robotic cell in study) are nominal data that often do not reflect accurately the real robotic environment. In this context, we may be planning robot paths for a robotic scenario that does not actually exist, at least in its original configuration. Thus we are in the presence of a Partially Unknown Environment (PUE). These differences between the idealized robot environment (the virtual one) and the real robotic environment can have different origins: the unpredictable dynamic behaviour of the real environment after contact with the robot or other equipments, robotic arm deflection, errors from the robot calibration process [9-10], an incorrect mapping of data from the virtual to the real environment, the roughness of contact surfaces, poorly representative CAD models and the presence of foreigner objects in the work environment. It follows from this that in order to have total control over the OLP process the robot has to know in real-time the actual configuration of its surrounding environment. In this way, robotic systems must have some degree of autonomy to overcome this situation. This has been achieved by incorporating sensors into the robotic systems [11-16]. Important studies have been conducted in this area, for example, the incorporation of sensors to increase industrial robot autonomy for welding applications [11-12] or for a general purpose robotic framework [13]. Sensor integration in task-level programming has also been a matter of study [14]. A number of vision-based solutions have been proposed to face PUEs. Kenney et al. use a vision-based approach to facilitate human-robot interaction and robot operation in unstructured environments [15]. Lopez-Juarez et al. explores force feedback to adapt robot behaviours to changing environments [16].

This paper proposes a force/motion control system to increase robot autonomy and thus to achieve a suitable robot performance in a PUE. The idea behind this is to control the end-effector pose (position and orientation) in real-time and in accordance with the forces and torques from the contact of the robot end-effector with its surrounding environment. This allows the robot to keep a given contact force and avoid undesirable impacts. The proposed force/motion control system has an external control loop based on forces and torques being exerted on the robot end-effector and an internal control loop based on robot motion. The external control loop is tested with a Proportional Integrative (PI) and a fuzzy-PI controller. The system performance is validated with real-world experiments involving contact in PUEs. Finally, results are discussed and some considerations about future work directions are made.

## 2. Force control applied to robotics

Over the last years, force control applied to robotics has assumed a growing importance in the proper execution of some robotic tasks [17]. These tasks are those in which the robot is required to maintain a given set force (deburring, polishing and assembly tasks) or others in which the deflexion of the robotic arm is a major factor (milling, grinding, drilling and friction stir welding). Even though these two cases appear to be different, both can be treated in the same way by applying a force control technique, passive force control [18] or active force control [18-23]. Hybrid force/motion control has been presented in literature as one of the most suitable methods to deal with PUEs [18].

Considering the approach proposed in this paper, this method allows controlling the non-constrained task directions (end-effector motion directions) in motion control and the constrained task directions in force control. The system is designed so that force control prevails over motion control. This means that position errors are tolerated to ensure force regulation.

Several robotic solutions using force control techniques have been developed and successfully applied to various industrial processes such as polishing [19] and deburring [20]. A number of force control techniques (fuzzy, PI, PID, hybrid, etc.) with varying complexity have been proposed thus far [20-23].

Fuzzy control was first introduced and implemented in the early 1970's in an attempt to design controllers for systems structurally difficult to model due to naturally existing nonlinearities and other modelling complexities [24]. Hsieh et al. present an optimal predicted fuzzy-PI gain scheduling controller to control the constant turning force process with a fixed metal removal rate under various cutting conditions [25]. Mendes et al. present a hybrid solution exploring robot force/motion control and different modalities of discretization and fitting of nominal data [26-27]. Lopes at al. present a force-impedance controlled industrial robot [28]. Gudur and Dixit propose a study in which the roll force and roll torque in a cold flat rolling process are modelled using first order Takagi-Sugeno (T-S) fuzzy models [29]. Many other studies apply T-S fuzzy models [30-34].

## 3. Robot control system

### 3.1. Hybrid force/motion control

Let's consider a rigid robot (manipulator) of *n* links, the dynamic equation of motion in the joint space is:

$$\mathbf{M(q)\ddot{q}+C(q,\dot{q})+B(q,\dot{q})+G(q)=\tau} \quad (1)$$

Where $\boldsymbol{\tau} \in \mathfrak{R}^n$ is the vector of applied joint torques, $\mathbf{q} \in \mathfrak{R}^n$ is the vector of joint positions, $\mathbf{M} \in \mathfrak{R}^{n \times n}$ is the inertia matrix, $\mathbf{C} \in \mathfrak{R}^n$ is the vector of Coriolis and centrifugal torques, $\mathbf{B} \in \mathfrak{R}^n$ is the vector of torques due to the friction action on the robot joints and $\mathbf{G} \in \mathfrak{R}^n$ is the vector of gravitational torques. When there is an external force applied to the robot end-effector, the dynamic equation (1) becomes:

$$\mathbf{M(q)\ddot{q}+C(q,\dot{q})+B(q,\dot{q})+G(q)}+\boldsymbol{\tau}_e =\boldsymbol{\tau} \quad (2)$$

Where $\boldsymbol{\tau}_e \in \mathfrak{R}^n$ is the vector of forces/torques exerted on the environment by the robot end-effector expressed in the robot joint space. This vector can be defined as:

$$\boldsymbol{\tau}_e = \mathbf{J}^T \mathbf{f} \quad (3)$$

Where $\mathbf{J}^T \in \mathfrak{R}^{n \times n}$ is the transpose of the Jacobian matrix and $\mathbf{f} \in \mathfrak{R}^n$ is the vector of forces and torques exerted on the environment by the robot end-effector expressed in Cartesian space. Thus, (3) may be written as:

$$\boldsymbol{\tau}_e = \mathbf{J}^T \mathbf{K} \mathbf{J}^T \Delta \mathbf{u} \quad (4)$$

Where $\mathbf{K} \in \mathfrak{R}^{n \times n}$ is the matrix of stiffness coefficients and $\Delta \mathbf{u} \in \mathfrak{R}^n$ is the vector of correction of displacements and orientations in the Cartesian space.

Traditionally, force/motion control systems applied to a robot manipulator have some end-effector motion directions controlled in motion and others controlled in force. In the proposed hybrid controller, all directions (along *x*, *y* and *z*) are controlled in motion (internal control loop) and some directions in force (external control loop). Contact forces and torques (between the robot tool and the robot working environments) are acquired from a force/torque (F/T) sensor which is between the robot wrist and the tool.

The external control loop processes the acquired information using a fuzzy-PI control system and sends end-effector position/orientation displacement corrections ($\Delta \mathbf{u}$) to the internal control loop, Fig. 1. The pre-programmed robot paths (nominal data) are then adjusted through the direct control of the servomotors of the robot. In Fig. 1, $\mathbf{q'}_k$ is the vector of joint positions in an instant of time previous to the current time.

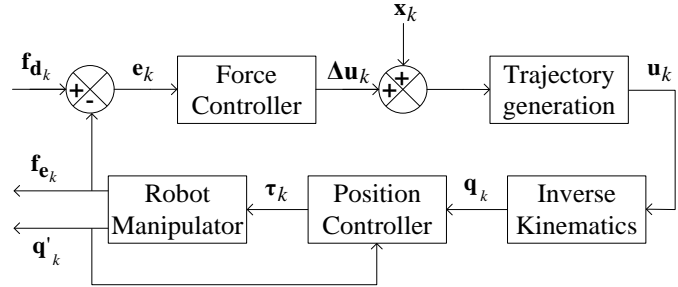

Fig. 1 Hybrid force/motion control system

## 3.2. Force controller

The proposed force controller associates PI control and fuzzy logic, a fuzzy logic controller type Mamdani [24]. The PI controller has good performance when applied in practical situations. A controller with derivative factor could help to decrease the correction time but it is very sensitive to noise. With regard to fuzzy, the controller type Mamdani is easy to implement and does not need a rigorous mathematical model of the system in study. Other types of fuzzy controller can require more rigorous mathematical models, for example the T-S controllers [29-34].

### 3.2.1. Fuzzy control architecture

The controller input variables are the force/torque error **e** and the change of the error **de**:

$$\mathbf{e}_k = \mathbf{f}_{\mathbf{d}_k} - \mathbf{f}_{\mathbf{e}_k} \tag{5}$$

$$\mathbf{de}_k = \mathbf{e}_k - \mathbf{e}_{k-1} \tag{6}$$

Where, $\mathbf{f_e}$ is the actual force/torque and $\mathbf{f_d}$ is the desired force/torque (set-points).

### 3.2.2. Fuzzy-PI

From the conventional PI control algorithm, the robot displacement **u** can be computed as:

$$\mathbf{u}(t) = \mathbf{K_P}\,\mathbf{e}(t) + \mathbf{K_I}\int \mathbf{e}(t)dt \tag{7}$$

Where $\mathbf{K_P}$ and $\mathbf{K_I}$ are coefficient constants. This can be represented in a discrete version:

$$\mathbf{u}_k = \mathbf{u}_{k-1} + \Delta\mathbf{u}_k \tag{8}$$

$$\Delta\mathbf{u}_k = \mathbf{K_P}\,\mathbf{de}_k + \mathbf{K_I}\,\mathbf{e}_k \tag{9}$$

If, $\mathbf{e}$ and $\mathbf{de}$ are fuzzy variables, (8) and (9) become a fuzzy control algorithm. A practical implementation of the proposed fuzzy-PI concept is in Fig. 2. Finally, the centre of area method was selected to defuzzify the output fuzzy set inferred by the controller:

$$\Delta U = \frac{\sum_{i=1}^{n} \mu_i\,\Delta U_i}{\sum_{i=1}^{n} \mu_i} \tag{10}$$

Where $\mu_i$ is the membership function which takes values in the range [0, 1]. A decision maker $\mathbf{S}$, $\mathbf{S} \in \Re^{n \times n}$, establishes the end-effector directions to control. When all directions are controlled in force and motion, the $\mathbf{S}$ matrix becomes the identity matrix.

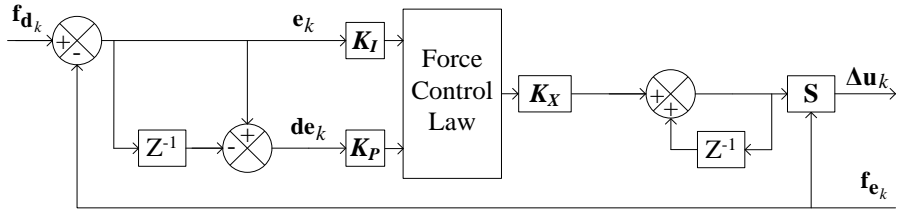

Fig. 2 Fuzzy-based force controller

### 3.2.3. Knowledge base

Each control variable is normalized into seven linguistic labels: positive large (PL), positive medium (PM), positive small (PS), zero (ZR), negative large (NL), negative medium (NM) and negative small (NS). The grade of each label is described by a fuzzy set. The membership function is in Fig. 3. The well known PI-like fuzzy rule base suggested by MacVicar-Whelan [35] is applied in this study, Table I.

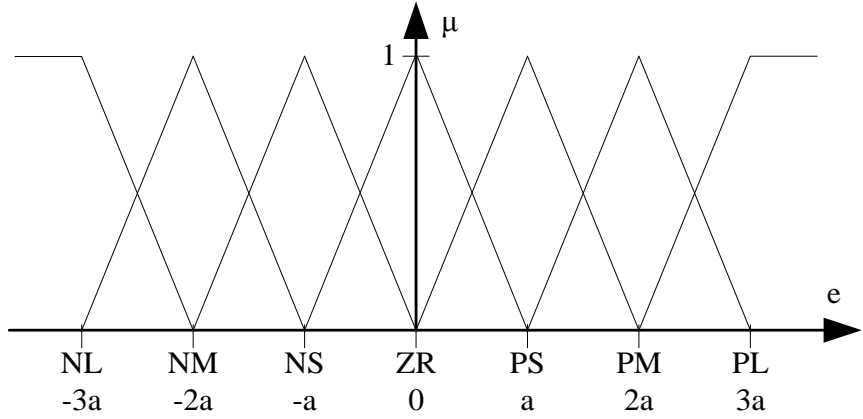

Fig. 3 Membership functions for the input variables

Table I: Representation of the rule base

| e\de | NL | NM | NS | ZR | PS | PM | PL |
|---|---|---|---|---|---|---|---|
| PL | nl | nm | ns | zr | pm | pl | pl |
| PM | nl | nl | nm | zr | pm | pl | pl |
| PS | nl | nl | ns | zr | ps | pl | pl |
| ZR | nl | nm | ns | zr | ps | pm | pl |
| NS | nl | nl | ns | zr | ps | pl | pl |
| NM | nl | nl | nm | zr | pm | pl | pl |
| NL | nl | nl | nm | zr | ps | pm | pl |

*3.2.4. Tuning strategy*

The system can be adjusted to different contact conditions by tuning the scaling factors $K_P$, $K_I$ and $K_X$ according to the characteristics of the environment in study. Lin and Huang propose an adjustment where the scaling factors are dynamic and thus they are adjusted at the same time the task occurs [21]. Also, different tables of rules can be used accordingly the task to be performed and the materials in contact involved in the task [20]. In this paper, the scaling factors are set to appropriate constant values, achieved by trial and error.

## 4. Experiments

The effectiveness of the proposed approach was evaluated in two real-world experiments involving contact in PUEs. In both experiments, the robot is programmed with nominal data from CAD drawings and the external control loop is tested with a PI and a fuzzy-PI controller, Fig. 4.

In the first experiment, the robot is programmed to manipulate plastic cups. A "foreign" object (a hammer) is introduced into the robot working environment, forcing it to become a PUE. This means that the nominal paths will drive the robot end-effector (with the plastic cup attached) to collide with the hammer. In this situation, when contact between the plastic cup and the hammer begins, the force/motion control system assumes the robot control, adjusting the end-effector to the PUE and maintaining a given value of contact force (10 N) along the $z$ axis and 0 N along the $x$ axis.

In the second experiment, the robot is programmed to be moved from a point to another in a straight path and maintaining contact with the workpiece, Fig. 6. In practice, since the contact surface of the workpiece is irregular and there are always calibration errors, it is impossible to properly perform the task described above without force/motion control. The force/motion control system assumes the robot control maintaining a given value of contact force (30 N) along the $z$ axis.

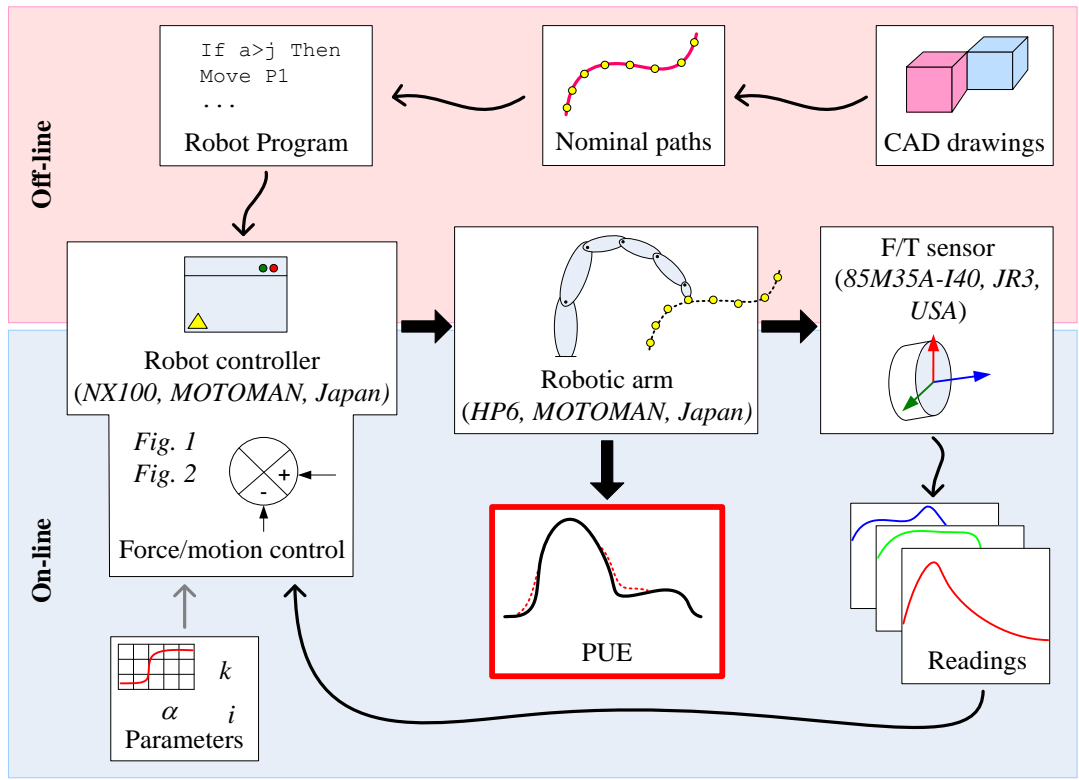

Fig. 4 Overview of the proposed approach and equipments

## 4.1. Results and discussion

For the first experiment, all tests showed similar force control results to those shown in Fig. 7 using a fuzzy-PI controller and those shown in Fig. 8 using a PI controller. Both systems provide acceptable results since the robot adapts to the PUE avoiding excessive contact forces. However, the fuzzy-PI controller performs better than the PI controller because the latter has a large overshoot and needs more time to stabilize, Fig. 8.

Results for the second experiment are shown in Fig. 9 when using a fuzzy-PI controller and in Fig. 10 when using a PI controller. The fluctuation in the controlled forces along the $z$ axis is due to the roughness of the contact surface. Nevertheless, these forces are all around the set point, 30 N. The PI controller has a better resolution (for small disturbances) than the fuzzy-PI controller. On the other hand, it presents a greater overshoot at the beginning of the convergence to the set point, Fig. 10. Since both systems (controllers) have similar results, a third experiment was done to ascertain the best solution. This experiment is similar to the second experiment but applying force/motion control along the $x$ axis and the $z$ axis, with set point forces of 6 N and 30 N, respectively. Results are in Fig. 11 and Fig. 12. In this context, results obtained along the $z$ axis are similar to those in the second experiment. For the forces controlled along the $x$ axis, it can be stated that the controller behaved the same way as for the control along the $z$ axis. Summarizing, we cannot say that one controller is better than the other.

The results obtained are in line with similar studies in the field that apply fuzzy reasoning to solve force/motion control problems [20-23].

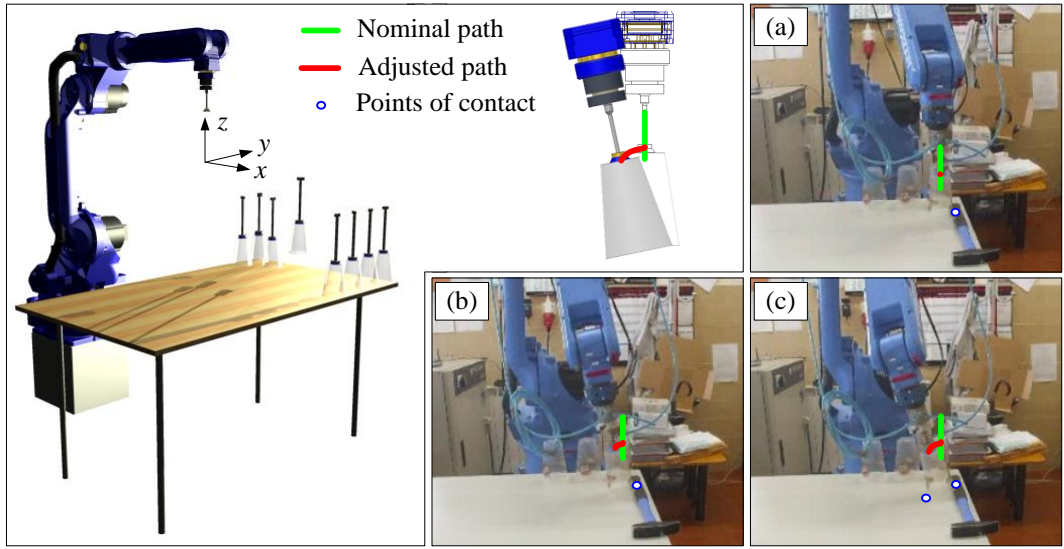

Fig. 5 Layout of the first experiment

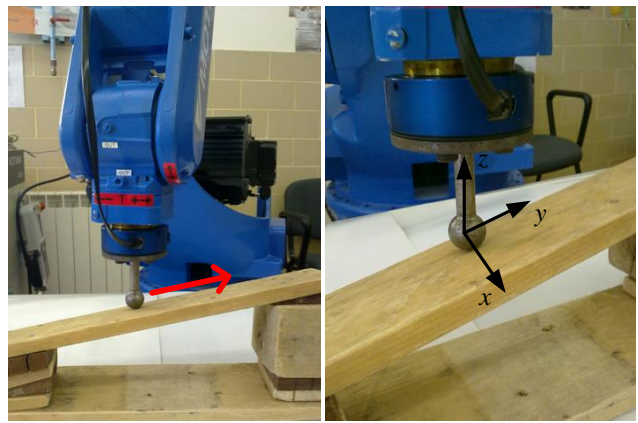

Fig. 6 Layout of the second experiment

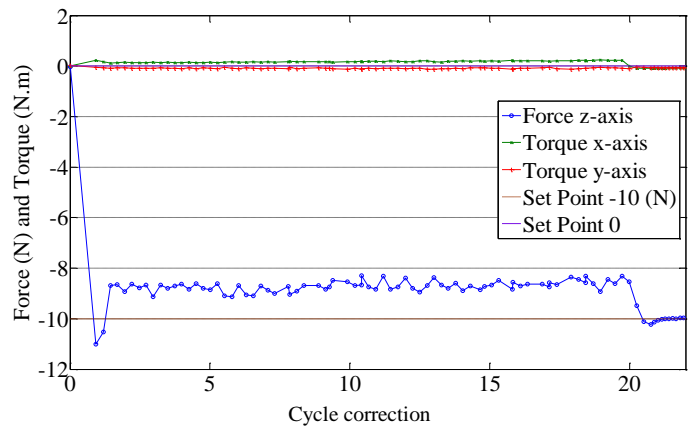

Fig. 7 Results for the first experiment using a fuzzy-PI controller

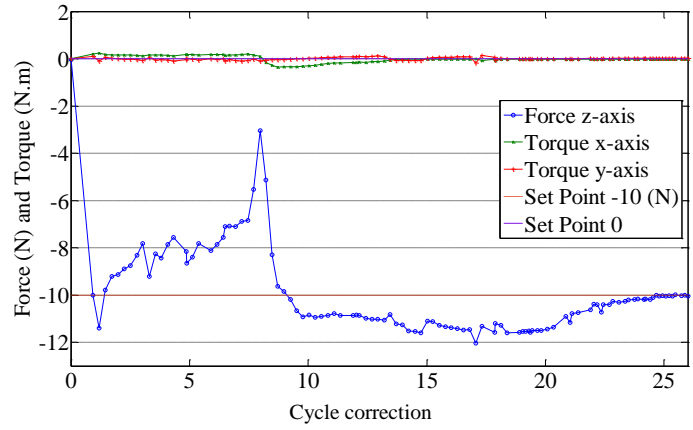

Fig. 8 Results for the first experiment using a PI controller

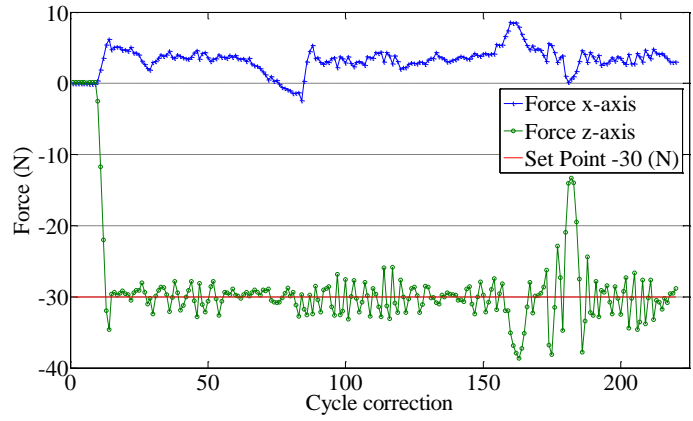

Fig. 9 Results for the second experiment using a fuzzy-PI controller

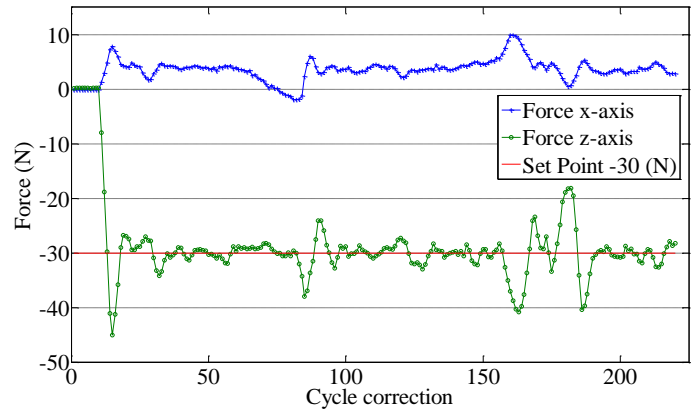

Fig. 10 Results for the second experiment using a PI controller

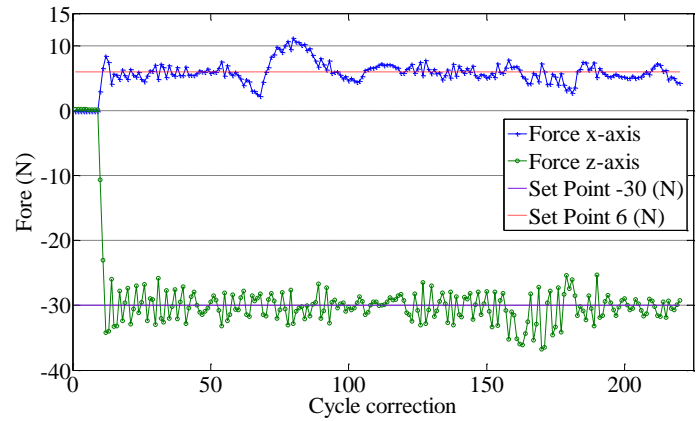

Fig. 11 Results for the third experiment using a fuzzy-PI controller

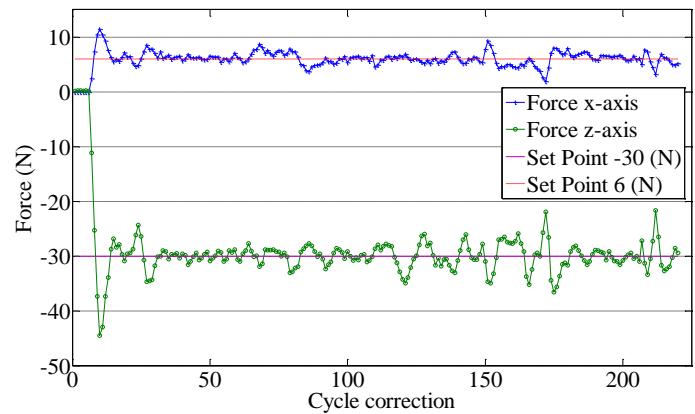

Fig. 12 Results for the third experiment using a PI controller

## 5. Conclusions

This paper presented a hybrid force/motion control system to increase robot autonomy. The system proved to be a valuable tool to help robots to adapt to PUEs, especially when contact exists. The external control loop of the hybrid controller was tested with a PI and a fuzzy-PI controller. Real-world experiments involving contact in PUEs demonstrated that we can not say that the fuzzy-PI controller is better than the PI controller. Both showed similar behaviours, with some disturbance around the set points. Another conclusion that can be drawn from experiments is that the proposed system only works properly if the data transfer between the F/T sensor and the robot controller is done in real-time. Future work will focus on performing more real-world experiments with different materials in contact.